\pdfoutput=1

\documentclass[11pt]{article}

\usepackage[]{acl}

\usepackage{times}
\usepackage{latexsym}
\usepackage{amsmath}
\usepackage{amssymb}
\usepackage{graphicx}

\usepackage{multirow}
\usepackage{url}
\usepackage{algorithm}
\usepackage{algpseudocode}
\usepackage{booktabs}
\usepackage{verbatim}

\usepackage[T1]{fontenc}

\usepackage[utf8]{inputenc}

\usepackage{microtype}

\title{Model-Based Simulation for Optimising Smart Reply}

\author{Benjamin Towle\textsuperscript{1}, Ke Zhou\textsuperscript{1,2} \\
  \textsuperscript{1}University of Nottingham \\
  \textsuperscript{2}Nokia Bell Labs \\
  \texttt{\{benjamin.towle, ke.zhou\}@nottingham.ac.uk} \\\
}

\begin{document}
\maketitle
\begin{abstract}
Smart Reply (SR) systems present a user with a set of replies, of which one can be selected in place of having to type out a response. To perform well at this task, a system should be able to effectively present the user with a diverse set of options, to maximise the chance that at least one of them conveys the user's desired response. This is a significant challenge, due to the lack of datasets containing sets of responses to learn from. Resultantly, previous work has focused largely on post-hoc diversification, rather than explicitly learning to predict sets of responses. Motivated by this problem, we present a novel method \textsc{SimSR}, that employs model-based simulation to discover high-value response sets, through simulating possible user responses with a learned world model. Unlike previous approaches, this allows our method to directly optimise the end-goal of SR--maximising the relevance of at least one of the predicted replies. Empirically on two public datasets, when compared to SoTA baselines, our method achieves up to 21\% and 18\% improvement in ROUGE score and Self-ROUGE score respectively. \footnote{This paper has been accepted to appear at ACL 2023.} 
\end{abstract}

\section{Introduction}
\begin{figure*}
    \centering
    \includegraphics[scale=0.12]{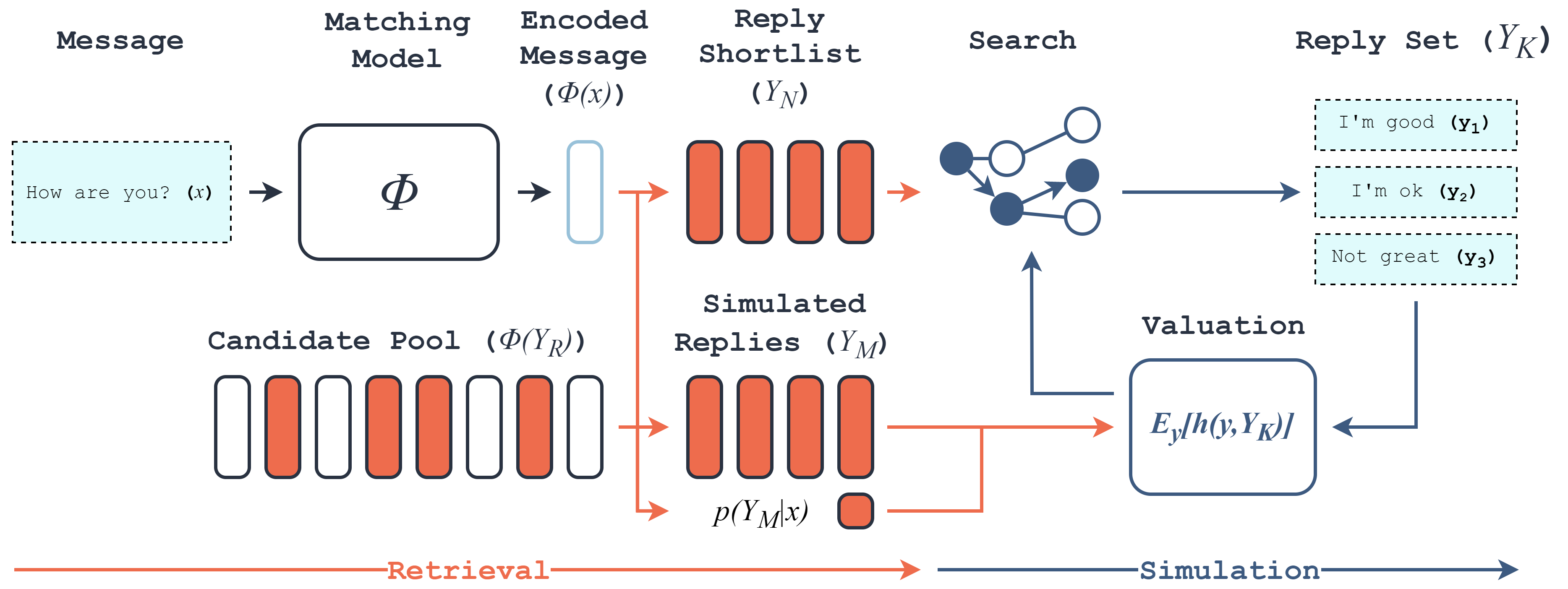}
    \caption{Overview of our approach. We combine a retrieval stage, which obtains the initial reply shortlist $Y_N$, followed by a simulation stage, which iteratively searches for reply sets $Y_K$ from that shortlist, and evaluates their relevance against a set of simulated replies $Y_M$.}
    \label{fig:simsr}
\end{figure*}


%

Automated response suggestion, or Smart Reply (SR), is rapidly becoming a staple feature of many email and chat systems such as Gmail, Skype, Outlook, Microsoft Teams, LinkedIn and Facebook Messenger. Given a message, SR systems present the user with a selection of possible responses, e.g. \texttt{How are you?} $\rightarrow \{$\texttt{I'm good}; \texttt{I'm ok}; \texttt{Not great}$\}$, which they can click in place of having to type out a reply. With the growth of communication over smaller devices that are poorly suited for manual typing \citep{Varcholik2012EstablishingAB, Palin2019HowDP}, such as smartphones and smart watches, SR is becoming an increasingly more important feature.

While early methods in SR incorporated sequence-to-sequence models \citep{Kannan2016SmartRA}, the current mainstream approach favours \emph{Matching models} which separately encode the message and reply into a shared latent space and retrieve the nearest neighbour response \citep{Deb2019DiversifyingRS, Zhang2021ADA, Deb2021ACG}. This has advantages in a production context, as it enables the model to retrieve replies from a fixed response set, maintaining greater controllability of model outputs; further, the latent representations for the response set can be pre-computed prior to inference, enabling faster latency. 

However, the naive approach of simply retrieving top-$K$ highest-scoring candidates from the Matching model often fails to produce a sufficiently diverse set of reply options. For instance, in response to the message \texttt{How are you?}, if the first predicted response is \texttt{I'm good}, predicting \texttt{I'm doing well} as the second response provides limited incremental value, as it carries equivalent semantic meaning. By contrast, \texttt{Not great} would be more useful, as it captures an alternative semantic meaning a user might wish to convey. In summary, one must account for the \textit{interdependencies} between replies. Previous methods have sought to implicitly account for these interdependencies such as through clustering by intent/topic, learning latent variables or re-scoring replies to include inter-reply similarity \citep{Kannan2016SmartRA, Deb2019DiversifyingRS, Deb2021ACG}. However, these techniques face two limitations: (1) they require hard-coded trade-offs between message-reply relevance and inter-reply diversity; (2) jointly optimising these two metrics is only partially correlated with the end goal of SR--maximising the relevance \textit{at least one} of the predictions. Ideally, it would be more principled if the model could simply optimise over this end goal. In so doing, we hypothesise performance would improve, while a good amount of diversity should also naturally emerge, insofar as it is correlated with performance on the task.

However, directly optimising this metric presents two problems: (1) the probability distribution over replies given messages is initially unknown; (2) we only have access to a \textit{single} reply for each message sampled from this distribution--i.e. the dataset of $\langle$message, reply$\rangle$ pairs--which prevents simply learning to predict reply sets via supervised learning. To circumvent these problems, we introduce model-based simulation (MBS) to the SR setting as a possible avenue forward. MBS is a technique from reinforcement learning \citep{Sutton2005ReinforcementLA} that allows an agent to choose what action to take by simulating the potential consequences of an action using a learned world model. We observe that the Matching model, given it is trained on a dataset of $\langle$message, reply$\rangle$ pairs, can also operate as a world model. This allows us to estimate the expected relevance of any reply set, by running repeated simulations with the world model. Crucially, relevance here can be defined as the maximum similarity between the reply set and a response sampled from the world model, which replaces the reliance on hard-coded trade-offs between message-reply relevance and inter-reply similarity.

Concretely, our method--\textsc{SimSR} (Figure \ref{fig:simsr})--comprises an initial retrieval stage, followed by an iterative simulation stage. We first retrieve a shortlist of replies from a larger candidate pool, using a learned neural Matching model, conditioned on a given message. In parallel, we also retrieve a number of simulated replies using the same method. Next, for the simulation stage, we use a search module to select a reply set comprising three responses from the shortlist. Then, we use a valuation module, which computes the expected similarity between the simulated replies and the most similar response from the reply set. This can be computed through a simple marginalisation process, using the probabilities and corresponding simulated replies provided by the world model. This process of search and valuation is iterated until the search algorithm terminates, and finally returns the highest scoring reply set. Quantitatively, our experiments show consistent out-performance against existing SoTA methods across two relevant datasets--Reddit and PERSONA-CHAT--achieving up to 21\% and 18\% improvement in ROUGE score and Self-ROUGE score respectively. \textsc{SimSR} also runs at a comparable speed to other methods, because the simulation is highly parallelisable and the Matching model only needs to encode the message once for both its initial retrieval and world model roles. In summary, our key contributions are:

\begin{itemize}
    \item We present model-based simulation as a novel paradigm for the Smart Reply task.
    \item We present \textsc{SimSR}, a novel method that employs model-based simulation with a learned world model.
    \item We demonstrate empirically the importance of taking into account reply interdependencies, achieving SoTA performance across the Reddit and PERSONA-CHAT datasets.
\end{itemize}

We make our code available for reproducibility.\footnote{\url{https://github.com/BenjaminTowle/SimSR}}

\section{Related Work}
\paragraph{Smart Reply.} In industry, SR has a range of applications from email systems to instant messaging. Naturally, the data from these is not publicly available to train on. Instead, recent work has made use of publicly available dialogue datasets such as Reddit \cite{Deb2021ACG, Zhang2021ADA}, which is sufficiently similar given SR applications are principally concerned with dialogue. While the earliest SR systems used sequence-to-sequence models \cite{Kannan2016SmartRA}, nowadays retrieval methods prevail which select a response from a pre-defined pool of candidates \cite{Henderson2017EfficientNL}, i.e. Matching models. By itself however, the Matching model has no way to ensure that the chosen reply set is sufficiently diverse. One approach to this is to ensure that no two responses in the reply set share the same topic/intent \cite{Kannan2016SmartRA, linkedinSR, Weng2019OCCAS}. However, this becomes more difficult in an open-domain setting, where the range of topics/intents is difficult to pre-define. As a result, other approaches have focused on more fine-grained diversification through conditional variational autoencoder techniques, which learn topics/intents across a continuous latent space during training \citep{Zhao2017LearningDDCVAE, Deb2019DiversifyingRS}. Maximum marginal relevance, which re-weights responses according to how similar they are with one another, has also been shown to work well \citep{Carbonell1998TheUO, Deb2019DiversifyingRS}. Our method differs from these approaches in that they employ diversity in a post-hoc manner which does not directly optimise the end goal of SR--maximising the relevance of at least one of the predicted replies.

\paragraph{Simulation in NLP.} In board games such as Go and chess, a model can have access to a perfect simulator, allowing it to explore various counterfactual trajectories before deciding what action to take next \citep{Silver2017MasteringCA}. In user-facing NLP applications, this is rarely possible. Therefore, much work has focused on settings such as self-play, in which a model learns to become better at a task such as negotiating \citep{Lewis2017DealON} or even open-domain dialogue \citep{Li2016DeepRL} through interacting with another copy of itself (or a version with frozen weights). User simulators are especially prevalent in task-oriented dialogue, where the domain is narrower and it is therefore easier to anticipate user behaviour \citep{Li2016AUS}. A notable exception to the above cases is text-based games--scripted games involving interacting in a wholly text-based environment--which are typically trained with access to a perfect simulator, as the game engine allows for previous states to be restored \citep{Jang2021LANGUAGEAV}. Our work is closest in spirit to those works that perform dialogue rollouts to select the next utterance using a reply prediction model \citep{Lewis2017DealON, Li2016DeepRL}--i.e. the Matching model. However, in our case the rollouts only involve a single step look-ahead, while our action space is the set of possible reply sets, rather than individual utterances. Further, our method can be used out-of-the-box during inference, without any further retraining of the Matching model. So far as we are aware, our work is the first to apply this concept of simulation to the SR setting.   

\section{Framework}


\subsection{Task Definition}
Our task is to predict a set of \textit{K} replies $Y_K = \{y_k\}_{k=1}^K$ from a candidate pool $Y_R$ of size $R$, conditioned on a message $x$. While in an online setting, the aim might be to maximise click-through rate \citep{Deb2019DiversifyingRS}, in an offline setting this can be approximated as maximising the similarity function $f(y)$, given as the maximum similarity between $Y_K$ and the ground truth response $y$ \citep{Zhang2021ADA}:
\begin{equation}
\label{eq:f_y}
    f(y) = \max_k[\{\mathrm{sim}(y,y_k)\}_{k=1}^K]
\end{equation}

\subsection{Matching Model}
\label{sec:matching}
Following previous approaches, we use a Matching model as the backbone of our method \citep{Henderson2017EfficientNL, Zhang2021ADA}. This comprises two parallel pre-trained transformer encoders $\Phi$ (with shared weights) that \textit{separately} encode $x$ and $y$ into a shared latent space. This is obtained by taking the output hidden-state corresponding to the \texttt{[CLS]} token which is pre-pended to each of the inputs. We refer to the vector representations of the message and reply as $\Phi(x)$ and $\Phi(y)$ respectively, and their score $g(x,y) = \Phi(x) \cdot \Phi(y)$. The model is trained using negative log-likelihood to maximise the joint probability of the context and reply:
\begin{equation}
\scriptstyle
p(x_i,y_i) = \frac{e^{g(x_i,y_i)}}{\sum_{y_j} e^{g(x_i,y_j)} + \sum_{x_j} e^{g(x_j,y_i)} - e^{g(x_i,y_i)}}
\end{equation}
This is referred to as \textit{symmetric loss} \citep{Deb2019DiversifyingRS}, and is known to impose tighter constraints on the relation between the message and reply, compared to having only a one-way classification loss function.

\section{SimSR}

For any given message $x$, there is uncertainty about the response $y$, which we assume to be sampled from some distribution $Y$. This is commonly referred to as the one-to-many problem \citep{Zhao2017LearningDDCVAE, towle-zhou-2022-learn} and is due to several reasons, such as unknown facts about the user and their intent. For example, the reply to \texttt{Can you meet for coffee at 2pm?} is likely to be conditioned on factors such as the user's schedule or their interest in meeting, which is unknown to a vanilla SR system. As a result, Matching models that simply select the most likely individual replies only achieve a lower bound of potential performance. This can be represented by the following inequality:
\begin{equation}
\label{eq:jensen}
E_{y \sim Y}[f(Y)] >= f(E_{y \sim Y}[Y])
\end{equation}
where $f(Y)$ refers to the similarity function from Equation \ref{eq:f_y}. The right hand side of Equation \ref{eq:jensen} represents what a Matching model approximates, while the left hand side is what we would like to obtain. Intuitively, this means that a good model should make predictions that capture the range of possible responses that could be sampled from $Y$, rather than simply the single most likely response. To do this, we hypothesise it is important to develop a method that accounts for the interdependencies between replies, i.e. which can evaluate sets of replies, rather than only individually scoring replies.

\begin{algorithm*}
\scriptsize
\caption{Model-Based Simulation with Ablative Search}
\hspace*{\algorithmicindent} \textbf{Input} Matching model $\Phi$, message $x$, response pool $Y_R$, number of candidates $N$, number of simulations $M$, final reply set size $K$. \\
\hspace*{\algorithmicindent} \textbf{Output} reply set $Y_K$ 
\begin{algorithmic}

\Procedure{ModelBasedSimulation}{$\Phi$, $x$, $Y_R$, $N$, $M$, $K$}
	\State $Y_N, Y_M, P_M  \gets \textproc{retrieve}(\Phi, x, Y_R, N, M)$ \Comment{Retrieve responses from $Y_R$ with corresponding probabilities $P_M$.}

	\State $C \gets \textproc{computeSimilarity}(Y_N, Y_M)$ \Comment{Obtain similarity matrix.}
	
	\While{$len(Y_N) > K$} \Comment{Ablative search loop}
		\State $L$, $bestScore$, $bestIdx  \gets \textproc{len}(Y_N)$, $-1.0$, $None$ 
		
		\For{$l \gets 0$ to $L$}
			\State $C_{tmp} \gets \textproc{concatenate}(C[:l],C[l+1:])$
			\State $eScore \gets \textproc{sum}(P_M \cdot \max(C_{tmp}, axis=0))$

			\If{$eScore > bestScore$} \Comment{Assign new best score if needed.}
				\State $bestScore, bestIdx  \gets eScore, l$
            \EndIf
        \EndFor
			
		\State delete $Y_N[bestIdx]$, $C[bestIdx]$ \Comment{Remove least useful reply}
    \EndWhile
    \State $Y_K \gets Y_N$
	\State \Return $Y_K$
 \EndProcedure

\end{algorithmic}
\label{alg:algorithm}
\end{algorithm*}

Algorithm \ref{alg:algorithm} and Figure \ref{fig:simsr} overview our method, which can be applied directly during inference. The Matching model first retrieves a shortlist of $N$ replies from a pool of pre-computed candidates $Y_R$ (Section \ref{sec:reply-shortlist}). Then we combine a search module which selects and constructs reply tuples from this shortlist to evaluate (Section \ref{sec:searchmethod}) and a valuation module (Section \ref{sec:valuation}) which computes an expected score between a given reply set and a list of simulated replies (Section \ref{sec:sim-replies}). Note that as our method does not require learning any new parameters, it can be applied to reply sets of arbitrary sizes during inference.

\subsection{Reply Shortlist}
\label{sec:reply-shortlist}
Given an overall candidate pool of size $R$, the corollary action space of $K$-tuples is intractably large: $\frac{R!}{K!(R-K)!}$. To mitigate this, we follow previous work \citep{Deb2019DiversifyingRS} and first retrieve the top-$N$ ranking replies conditioned on the message $x$, using the Matching model, where $N << R$. We refer to this set as $Y_N = \{y_n\}_{n=1}^N$. This defines the building blocks with which we can construct the action space of $K$-tuples of replies to perform our simulation on. 

\subsection{Simulated Replies}
\label{sec:sim-replies}

We do not have access to the ground-truth data-generating distribution--i.e. $p_{human}(y|x)$--which would be required for planning in the actual environment. However, the Matching model can serve as an effective approximator of this distribution--henceforth, $p_{model}(y|x)$--since it was trained on $\langle$message,reply$\rangle$ pairs sampled from the ground-truth distribution. Thus, using the same Matching model as above, we retrieve the top-$M$ replies, also conditioned on the message $x$, to obtain $Y_M = \{y_m\}_{m=1}^M$. In practice, as we use the same model to retrieve both $Y_N$ and $Y_M$, this can be achieved with a single query of the response set--therefore, the impact on latency is kept to a minimum.

\subsection{Valuation}
\label{sec:valuation}
We define similarity between a $K$-tuple and the $m$-th simulated response $y_m \in Y_M$ as:
\begin{equation}
h(y_m,Y_K) = \max_k \{\mathrm{sim}(y_m, y_k)\}_{k=1}^K
\end{equation}
where $\mathrm{sim}(\cdot,\cdot)$ is a similarity score. Intuitively, this rewards the model if at least one of the predictions is relevant to the user. We use term-level F1-score to represent similarity for simplicity, and leave alternative measures for future work. We obtain the expected similarity for a given $K$-tuple by marginalising over the scores for all $y_m \in Y_M$:
\begin{equation}
\label{eq}
    E[h(y, Y_k)] = \sum_m^M h(y_m,Y_K) \cdot p_{model}(y_m|x)
\end{equation}
In practice, we found dividing the scores by a high temperature ($\tau = 10$) \citep{Hinton2015DistillingTK} before applying a softmax normalisation improved performance, as it encouraged the model to take into account a larger range of possible simulated responses.

\subsection{Search}
\label{sec:searchmethod}
Given our method for estimating the value of any given $K$-tuple, it is necessary to employ a search algorithm, to decide which tuples should be evaluated. In this work, we consider a selection of out-of-the-box and bespoke methods:

\paragraph{Exhaustive Search.} A straightforward approach is to simply enumerate and evaluate all possible tuples. This is feasible because (a) $N$ is typically a relatively small number ($15$, in our experiments), (b) the computational cost for evaluating any given tuple is low, given it involves simply computing Equation \ref{eq} where the similarity function $\mathrm{sim}(\cdot,\cdot)$ only needs to be computed once for each $y_n$,$y_m$ pair.

\paragraph{Ablative Search.} For larger values of $N$, it is necessary to employ a more selective search strategy. We observe that the task of finding $K$ replies from a shortlist of $N$ replies can be treated partially as a clustering problem, where each reply in the $K$-tuple represents a cluster nucleoid, and the objective is to minimise some distance matrix.  To this extent, we design a method that incrementally builds the reply set by iteratively removing (hence, \textit{ablative}) the least useful reply from the shortlist $N$, until only $K$ replies remain. In detail, for each of the $(N-1)$-tuples of $Y_N$ we compute $E[h(y,Y_{N-1})]$, such that $Y_{N-1}^*$ is the $(N-1)$-tuple that obtained the highest score. We then remove the sole reply $y*$ from $Y_N$ that is not present in $Y_{N-1}^*$. Finally, we repeat this process for all of the $(N-2)$-tuples of $Y_{N-1}$ etc. until we are left with $Y_{N-(N-K)} = Y_K$.

\paragraph{Greedy Search.} A limitation of ablative search is that it requires a lot of non-parallelisable compute due to the iterative nature of the algorithm. We therefore consider a greedy alternative. In brief, instead of obtaining $Y_K$ by whittling down $Y_N$, we instead incrementally build up $Y_K$ starting from the empty set. This thus requires only $K$ non-parallelisable steps, rather than $N - K$. In detail, let $Y_G$ be the set of currently chosen replies, such that initially $Y_G = \varnothing$. Then, for each reply $y_n \in Y_N$ we compute the expected similarity for the union of $Y_G$ and $y_n$, i.e. $E[h(y, Y_G \cup y_n)]$. Next, we append the highest scoring $y_n$ to $Y_G$, and repeat until $|Y_G| = K$.

\paragraph{Sample and Rank.} Finally, we consider a simple sample and rank approach, which has been shown to work well in other NLP tasks such as dialogue \citep{meena}. This involves randomly selecting a subset of all possible tuples, and evaluating them. Then, we return the tuple with the highest score according to Equation \ref{eq}.

\section{Experiments}
\begin{table*}[]
\centering
\small
\begin{tabular}{lccccc}
\toprule
\multirow{2}{*}{Search}               & \multicolumn{2}{c}{Reddit}                                 & \multicolumn{2}{c}{PERSONA-CHAT} & \# Tuples                                                           \\
\cmidrule(lr){2-3} \cmidrule(lr){4-5}
                & ROUGE $\uparrow$ & Self-ROUGE $\downarrow$ & ROUGE $\uparrow$ & Self-ROUGE $\downarrow$ & Evaluated $\downarrow$ \\
                 \midrule
Exhaustive    & 2.47  & 2.49 & \textbf{7.85} & 8.60 & 455 \\
Ablative    & 2.40  & \textbf{2.36} & 7.71 & \textbf{8.39} & 114 \\
Greedy     & \textbf{2.49} & 2.77 & 7.82 & 9.76 & 42 \\
Sample-and-Rank    & 2.39 & 2.79 & 7.39 & 12.27 & \textbf{25} \\
\bottomrule
\end{tabular}
\caption{Results on the Reddit and PERSONA-CHAT Test sets under different search strategies for \textsc{SimSR}.}
\label{tab:search}
\end{table*}
We now turn our attention towards empirical testing of \textsc{SimSR}, addressing the following research questions:

\begin{itemize}
    \item \textbf{RQ1:} How does the choice of search strategy impact relevance and diversity in \textsc{SimSR}? (Section \ref{sec:search})
    \item \textbf{RQ2:} How does \textsc{SimSR} compare to existing SoTA SR methods? (Section \ref{sec:main_results}, \ref{sec:case_study})
    \item \textbf{RQ3:} How much does \textsc{SimSR} benefit from accounting for interdependencies between replies when selecting a reply set? (Section \ref{sec:ablation})
\end{itemize}

\subsection{Baselines}

We identify four types of diversification strategies which serve as baselines against our model. The original implementations of these methods are typically proprietary and unavailable for direct comparison. Therefore, in the list below we summarise our re-implementations as well as key changes that were made versus the original.

\paragraph{Matching} is the base retrieval model discussed earlier (Section \ref{sec:matching}) \citep{Henderson2017EfficientNL, Zhang2021ADA}. It simply selects the top-$K$ responses according to their individual scores without any additional components. Our version uses the DistilBERT model as a base \citep{Sanh2019DistilBERTAD}, whereas previous methods used a variety of transformers \citep{Zhang2021ADA} and recurrent neural networks \cite{Deb2019DiversifyingRS}--we follow this for all baselines.
\paragraph{Matching-Topic} uses topic classification to ensure none of the top-$K$ responses share the same topic \citep{Kannan2016SmartRA, linkedinSR, Weng2019OCCAS}. We replace the classifier with an out-of-the-box classifier trained on Twitter \citep{twittertopic}, which features similarly short-form messages to those used in SR. 
\paragraph{Maximum Marginal Relevance (MMR)} re-weights responses according to how similar they are with one another, which is combined in a linear combination with their message-response score \citep{Deb2019DiversifyingRS}. Our re-implementation is closer to the original algorithm \citep{Carbonell1998TheUO} in that we incrementally build the reply set, rather than in a single step--we found this performed better during early testing.
\paragraph{MCVAE} \citep{Deb2019DiversifyingRS} is a conditional variational autoencoder \cite{Zhao2017LearningDDCVAE} built on top of the Matching model, allowing for multiple query vectors to be generated from a single message embedding. Candidates are scored using a voting process whereby each query vector selects the nearest reply, and the $K$ most-selected replies are chosen. We re-implement this without any major changes from the original to the best of our knowledge, and use the original paper's hyperparameters, such as size of the latent variable, where possible.

\subsection{Datasets}
\begin{table}[]
\small
    \centering
    \begin{tabular}{lcccccc}
    \toprule
        & \multicolumn{3}{c}{Reddit} & \multicolumn{3}{c}{PERSONA-CHAT}  \\
        \cmidrule(lr){2-4} \cmidrule(lr){5-7}
         & Train & Valid & Test & Train & Valid & Test \\
         \midrule
         \# Samples & 50k & 5k & 5k & 66k & 8k & 8k \\
         \bottomrule
    \end{tabular}
    \caption{Statistics for the datasets.}
    \label{tab:datasets}
\end{table}
We evaluate our methods across two datasets, summarised in Table \ref{tab:datasets}. While most prior work has used proprietary datasets \citep{Kannan2016SmartRA, Deb2019DiversifyingRS}, we identify a single publicly available SR dataset--Reddit/MRS \citep{Zhang2021ADA}. We supplement this by also evaluating on PERSONA-CHAT \citep{Zhang2018PersonalizingDA}, which similarly falls under the broader umbrella of open-domain dialogue. Below we provide further elaboration:

\paragraph{Reddit} or MRS \citep{Zhang2021ADA} is, to the best of our knowledge, the only publicly available dataset created specifically for the SR setting. The dataset is multilingual, covering $10$ languages and over $50M$ message-reply pairs extracted from the social-media site Reddit. As our focus is only on the monolingual setting, we use only the English portion of the corpus. Further, due to limited computational resources we train and evaluate on only a small subset of the data (randomly selected). 
\paragraph{PERSONA-CHAT} \citep{Zhang2018PersonalizingDA} is a crowdworker-sourced dialogue dataset between pairs of speakers in which each speaker is assigned a brief persona comprising a few sentences, e.g. \texttt{I have a dog}. We simply concatenate this information to the message, following previous approaches \citep{Humeau2020PolyencodersAA}. As it is an open-domain dialogue dataset, it covers a broad range of possible conversations, and therefore provides another useful benchmark of performance for an SR system, which are often deployed in similarly open-domain environments.

\subsection{Metrics}
We use a weighted ROUGE \citep{Lin2004ROUGEAP} ensemble metric to evaluate performance, which is known to be well correlated with click-through rate in the SR setting \citep{Zhang2021ADA}. This consists of a mixture of 1/2/3-grams for ROUGE-F1:

\begin{equation}
    \frac{\textsc{rouge-1}}{6} + \frac{\textsc{rouge-2}}{3} + \frac{\textsc{rouge-3}}{2}
\end{equation}

\subsection{Hyperparameters}
We train our models using the Adam optimizer \citep{Kingma2014AdamAM} for $3$ epochs, with an initial learning rate of $5e-5$ and linear decay, and a batch size of $8$. We truncate the message and response to the last $64$ tokens each. We initialise our models from the DistilBERT checkpoint \citep{Sanh2019DistilBERTAD},\footnote{\url{https://huggingface.co/distilbert-base-uncased}} which is a $66M$ parameter transformer trained via knowledge distillation on BERT. During inference, we set $K = 3$ which is a standard number for SR \citep{Zhang2021ADA}. We also set the number of candidates initially retrieved by the Matching model $N = 15$, which previous work has shown provides a good trade-off between accuracy and latency \cite{Deb2019DiversifyingRS}. For \textsc{SimSR}, we set the number of simulations $M = 25$. For both PERSONA-CHAT and Reddit we use the entire training set to retrieve from (i.e. $Y_R$). In early testing, we explored using heuristic techniques to create a more deduplicated candidate pool, but found limited benefit, and therefore opted for this simpler approach.

During deployment, although SR systems produce multiple replies, only \textit{one} of them needs to be relevant. To replicate this, we only record the maximum ROUGE across the $K=3$ replies outputted. We also report Self-ROUGE \citep{Celikyilmaz2020EvaluationOT}, which is an unreferenced metric that measures the diversity of the predicted replies. For each reply $y_k \in Y_K$, we treat $y_k$ as the prediction and the other two replies as the references, using the same ROUGE metric as above. Note that a lower Self-ROUGE indicates \textit{more} diversity. 

\subsection{Choosing a Search Strategy}
\label{sec:search}

Table \ref{tab:search} shows the performance of \textsc{SimSR} under different search strategies. This is motivated by two sub-questions: (1) how robust is \textsc{SimSR} to the choice of search strategy? (2) What trade-offs are involved between relevance, diversity and efficiency?  

Exhaustive search unsurprisingly performs the best both in terms of relevance and diversity, but is the least efficient and would not scale to larger values of $N$. More interesting is the trade-off between relevance and diversity that occurs between the Ablative and Greedy methods. Greedy performs slightly better in relevance, perhaps suggesting that the longer sequences involved in the Ablative method leave more opportunity for errors to be propagated. However, Greedy performs significantly worse in diversity. While a high diversity is not always a good thing (e.g. random guessing would also have a high diversity), Ablative's diversity is much closer to that obtained by Exhaustive search. Sample and Rank consistently gave the worst results, suggesting randomly constructing tuples is insufficient for finding high-value tuples.

Overall, these results show that \textsc{SimSR} is reasonably robust to the choice of search strategy. Going forward, we opt to use Ablative search for subsequent experiments which provided arguably the best trade-off in terms of relevance, diversity and efficiency by a small margin.

\subsection{Main Results}
\label{sec:main_results}

\begin{table*}[]
\centering
\small
\begin{tabular}{llcccc}
\toprule
\multirow{2}{*}{Section} & \multirow{2}{*}{Method}                & \multicolumn{2}{c}{Reddit}                                 & \multicolumn{2}{c}{PERSONA-CHAT}                                                           \\
\cmidrule(lr){3-4} \cmidrule(lr){5-6}
                 & & ROUGE $\uparrow$ & Self-ROUGE $\downarrow$ & ROUGE $\uparrow$ & Self-ROUGE $\downarrow$ \\
\midrule
\multirow{4}{*}{(A) Baselines} & Matching         & 2.04                           & 6.92                                &  6.61                           & 12.44 \\                                

& Matching + Topic & 2.01                           & 3.17                               & 6.42                           & 11.77  \\                              
& Matching + MMR & 2.17                           & 5.19                                & 6.66                           & 10.76  \\ 

& MCVAE            & 2.12                           & 3.99                                & 6.52                           & 8.93 \\                               
\midrule
(B) Our Method & \textsc{SimSR}       & \textbf{2.40}                           & \textbf{2.36}                               & \textbf{7.71}                           & \textbf{8.39}     \\
\midrule
\multirow{2}{*}{(C) Ablations} & - Multi-reply & 2.02 & 19.77 & 7.03 & 35.24 \\
& - Simulation & 2.04                           & 6.92                                &  6.61                           & 12.44 \\
\bottomrule
\end{tabular}
\caption{Performance of \textsc{SimSR} (B) compared to baseline approaches (A) and ablations (C) on the Reddit and PERSONA-CHAT Test sets. All results are statistically significant on t-test with \textit{p}-value < 0.01.}
\label{tab:main_results}
\end{table*}

Table \ref{tab:main_results}A-B summarises our main results. Across both tasks, we find that additional filtering/diversification measures improve the diversity of the suggested replies, but provide only limited improvement to relevancy. We argue this reflects the fact the these methods often involve trading off relevance for diversity, such as MMR, which explicitly scores replies as a linear combination of their relevancy to the message and their similarity to other replies in the reply set. Similarly, whilst the out-of-the-box Topic classifier sometimes produced outputs that were more diverse than the other baselines, this came at the cost of reduced relevance, due to it being too coarse-grained--i.e. often a given message required multiple replies from the \textit{same} topic.

Contrastingly, we show our method is able to consistently improve on both relevancy and diversity for both tasks. On Reddit, relevancy improves by up to 14\% and diversity by up to 21\%; on PERSONA-CHAT, relevancy improves by 18\% and diversity improves by 6\%. All results are statistically significant on a t-test with \textit{p}-value < 0.01. The main difference between the datasets is that PERSONA-CHAT is a less noisy dataset, being made by crowdworkers, and therefore both metrics are comparatively higher.

\subsection{Ablations}
\label{sec:ablation}
We consider the question of whether \textsc{SimSR} is simply learning to predict individual replies that have a high expected score, rather than learning to take advantage of interdependencies between replies. To this end, in Table \ref{tab:main_results}C we present an ablation (`- Multi-Reply') that selects the top-$K$ replies according to their \textit{individual} scores in simulation, without considering their scores at the \textit{tuple}-level, i.e. $\mathrm{TopK}(\{E[h(y,y_n)]\}_{n=1}^N)$. We also present a version without simulation at all as a baseline comparison, which is equivalent to the Matching model in Table \ref{tab:main_results}A. 

Results show that removing multi-reply significantly harms performance. Versus the baseline, there is no improvement on Reddit, while there are only limited gains on PERSONA-CHAT, suggesting most of the performance gains from \textsc{SimSR} are due to the ability to account for interdependencies within the reply set. We hypothesise the reason for the difference between the two datasets is because PERSONA-CHAT is a less noisy dataset, and therefore selecting individual replies with a high expected similarity may provide some benefit. Diversity is especially harmed, and even is significantly less diverse than the baseline. This is unsurprising, given maximising the similarity of each reply to the same set of simulated replies implicitly encourages responses to be similar.

\subsection{Case Study}
\label{sec:case_study}

\begin{table*}[]
    \centering
    \small
    \begin{tabular}{l|l}
    \toprule
    \multicolumn{1}{c|}{\textbf{PERSONA-CHAT}} & \multicolumn{1}{c}{\textbf{Reddit}} \\
    \midrule
    \textbf{Message: } \textit{So do you have any pets?} & \textbf{Message: } \textit{where? i've always wanted to be in one!} \\
    \midrule
        \multicolumn{2}{c}{\textbf{Matching}} \\
        \midrule
        No, no pets. Do you have any & I'm so glad I'm not the only one. \\
        No, no pets. You? & glad i'm not the only one....\\
        No, I do not have any pets. What are some things you like & Wait... They said I'll be the the first... \\
        \midrule
        \multicolumn{2}{c}{\textbf{MMR}} \\
        \midrule
        I do not have any but I do want a dog & I will have one of everything, please. \\
        No, no pets. You? & I'm so glad I'm not the only one. \\
        No, no pets. Do you have any? & glad i'm not the only one.... \\
        \midrule
        \multicolumn{2}{c}{\textbf{\textsc{SimSR}}} \\
        \midrule
        No, I do not have any pets. & I'll be there, too. Also my first time seeing them. Can't wait. \\
        Nope no pets at the moment. How are you? & Glad I wasn't the only one \\
        Yes I have 2 dogs. & ME TOO. We need to go find one. \\
        \bottomrule

    \end{tabular}
    \caption{Examples of model outputs on the PERSONA-CHAT (left) and Reddit (right) Test sets. \textsc{SimSR} produces replies that capture multiple possible user intents, while the other approaches capture a more limited range of intents.}
    \label{tab:case_study}
\end{table*}

Table \ref{tab:case_study} presents two case studies comparing the qualitative performance of \textsc{SimSR} versus a selection of baseline methods. In both case studies we see \textsc{SimSR} is able to represent three diverse intents across its predictions versus only one or two intents for the Matching and MMR models. In the left example, \textsc{SimSR} is crucially able to capture including both a positive and a negative intent, unlike the baselines. In the right example, \textsc{SimSR} successfully avoids duplicating the \texttt{I'm glad} intent. Note that in both cases it would be impractical to use heuristic measures to deduplicate the intents (e.g. removing replies with only 1 word edit distance) as there is often only partial term-level overlap between the utterances.

\subsection{Latency}
\begin{table}[]
\small
    \centering
    \begin{tabular}{lc}
    \toprule
        Method & Latency (ms) \\
        \midrule
        Matching & 23.3  \\
        Matching + Topic & 45.5 \\
        Matching + MMR & 24.5 \\
        MCVAE & 25.9 \\
        \midrule
        \textsc{SimSR} & 29.9 \\
        \bottomrule
    \end{tabular}
    \caption{Latency of \textsc{SimSR} compared to baseline approaches on the Reddit Validation set.}
    \label{tab:latency}
\end{table}
Table \ref{tab:latency} validates the limited latency impact of \textsc{SimSR} compared to the baseline methods. We used an NVIDIA GeForce RTX 3060 Ti GPU and CPU operations were conducted by an AMD Ryzen 7 5700G with Radeon Graphics. For the initial retrieval, we pre-compute the reply embeddings and store them in a FAISS index \citep{faiss}. Overall, we find \textsc{SimSR} is able to maintain comparable latency to other methods which encorporate post-hoc diversification methods such as MCVAE and MMR. The small latency difference for \textsc{SimSR} is mainly due to the iterative search and evaluation process not using any low-level optimisation in the code or multiprocessing. Topic is the slowest due to the additional inference cost of the Topic classifier.

\section{Conclusion}
In this work, we have presented a method for generating sets of replies for Smart Reply systems, using model-based simulation and a range of search strategies to discover high-value reply sets, without the need for any additional training. Our method outperforms existing SoTA methods on both datasets tested, and we have supported our results by detailed analysis of the effect of different search strategies, demonstration of the importance of accounting for interdependencies between replies, and a detailed case study. Future work could consider whether it is possible to improve the quality of the initial retrieval (e.g. by training on sets of replies), or other methods for scoring response similarity during simulation.

\section*{Acknowledgements}
We thank the reviewers for their helpful feedback and suggestions during the reviewing process. This work is partly supported by the EPSRC DTP Studentship program. The opinions expressed in this paper are those of the authors, and are not necessarily shared or endorsed by their employers and/or sponsors.  

\section*{Limitations}
While our approach is able to optimise over the retrieved shortlist of replies, it does not improve the initial retrieval from the candidate pool, which still scores individual candidates, rather than reply sets, using the Matching model. This is a limitation that is shared with prior baseline methods. A further limitation is that we only consider the monolingual setting, whereas many deployed SR applications have an international footprint. Learning a multilingual Matching model in SR is known to have additional challenges \citep{Deb2021ACG}. Another limitation is that our model is only tested on public dialogue datasets, due to actual conversations on platforms using SR being proprietary. Therefore, while our techniques should work well in the instant messaging setting, our methods have not been directly tested in the email setting.

\section*{Ethical Considerations}
As neural dialogue models have grown in expressive capabilities and fluency, ethical considerations are an increasingly prominent issue. Key considerations typically centre around model's tendencies (1) to produce information that is factually inaccurate \citep{Shuster2021RetrievalAR} or (2) to repeat toxic/biased behaviour from the training data \citep{Xu2020RecipesFS}. Compared to vanilla dialogue models, these risks are mitigated in SR: (1) SR is usually limited to short-form replies that express simple information, and is therefore less likely to lead to the kinds of hallucination seen in longer-form answers; (2) SR typically does not generate tokens sequentially, but retrieves responses from a pool of candidates, which can be vetted in advance. Note however, this does not prevent replies that are contextually inappropriate when paired with a particular message, e.g. \texttt{Do you hate people?} $\rightarrow$ \texttt{Yes, I do}. The human-in-the-loop, who must ultimately choose and be accountable for whether or not to select one of the suggested replies, can be seen as a risk mitigant compared to vanilla chatbots. Conversely however, \citet{Wenker2022WhoWT} identify risks pertaining to a loss of human agency, such as due to a user selecting a sub-optimal reply to save time or being primed by the replies. This could lead to people being more trusting of an SR-generated reply versus receiving a reply from a chatbot, due to the belief that a human ultimately is behind it. We also only experimented with datasets that were released by previous studies, which are publicly available. These datasets (especially Reddit) often contain toxic/biased behaviour which developers should bear in mind if using this system in a deployment context.

\bibliography{anthology,custom, local}
\bibliographystyle{acl_natbib}

\appendix
\section{Artifacts: code, datasets and models}
This section lists the licences for the code, datasets and models used in the paper (`Artifacts'): DistilBERT \citep{Sanh2019DistilBERTAD} is under Apache-2.0 licence; PERSONA-CHAT \citep{Zhang2018PersonalizingDA} is under CC BY 4.0; The topic classifier \citep{twittertopic} is fine-tuned from the pre-trained transformer RoBERTa \cite{Liu2019RoBERTaAR} which is under the MIT licence; The Reddit dataset \citep{Zhang2021ADA} is available under the MIT licence; Our code pertaining to this paper is released under the MIT licence.
\section{Experiment details}
Models were trained using an NVIDIA GeForce RTX 3060 Ti. Training took no longer than an hour for any one model, as they were fine-tuned from pre-existing pre-trained models and the datasets were comparably small. Hyperparameters were selected using using recommended values for fine-tuning \citep{Devlin2019BERTPO}, and where not explicitly specified use default values from the HuggingFace Trainer class. Experiments were run using a single random seed. For evaluation, ROUGE was calculated using the rouge-score Python package \footnote{\url{https://pypi.org/project/rouge-score/}}.

\section{Further examples}
Table \ref{tab:further_examples} shows further examples of \textsc{SimSR}'s predictions versus the other baselines.
\begin{table*}[]
    \centering
    \small
    \begin{tabular}{ll}
    \toprule
    \multicolumn{2}{c}{PERSONA-CHAT} \\
    \midrule
     \textbf{Message:} & \textit{i do, i turn up ed sheeran on my ipod and go to my favorite waterfall.} \\
    \midrule

        \multirow{3}{*}{\textbf{Matching}} & that is nice do you like to hike ? \\
        & do you like to hike ?\\
        & that sounds like fun . do you have a favorite artist ? \\
        \midrule
        \multirow{3}{*}{\textbf{MMR}} & who is your favorite artist ? \\
        & that is nice do you like to hike ? \\
        & do you like to hike ? \\
        \midrule
        \multirow{3}{*}{\textbf{\textsc{SimSR}}} & do you like to hike ? \\
        & that is amazing . i love nature . \\
        & who is your favorite artist ? \\
        \midrule
    \multicolumn{2}{c}{Reddit} \\
    \midrule
    \textbf{Message:} & \textit{deal. i'm in newcastle haha} \\
    \midrule
        
        \multirow{3}{*}{\textbf{Matching}} & See you there! \\
        & Great, see you there! \\
        & I'm In \\
        \midrule
        \multirow{3}{*}{\textbf{MMR}} & Where? I’m low on gas and you need a jacket. \\
        & See you there! \\
        & Great, see you there! \\
        \midrule
        \multirow{3}{*}{\textbf{\textsc{SimSR}}} & see you in 15 minutes. \\
        & Yeah sure \\
        & Sounds good to me tho \\
        \bottomrule

    \end{tabular}
    \caption{Additional examples of model outputs on the PERSONA-CHAT (top) and Reddit (bottom) Test sets.}
    \label{tab:further_examples}
\end{table*}



\end{document}